\pdfoutput=1

\documentclass[11pt]{article}

\usepackage[]{EACL2023}

\usepackage{times}
\usepackage{latexsym}
\usepackage{svg}
\usepackage{amsmath,bm}
\usepackage{amssymb}
\usepackage[T1]{fontenc}

\usepackage[utf8]{inputenc}
\usepackage{caption}
\usepackage{subcaption}
\usepackage{multirow}
\usepackage{enumitem}
\usepackage{lipsum}
\usepackage{array}
\usepackage{microtype}

\usepackage{inconsolata}

\title{Span-based Named Entity Recognition \\by Generating and Compressing Information}

\author{Nhung T. H. Nguyen\textsuperscript{\textnormal{1}}, 
         Makoto Miwa\textsuperscript{\textnormal{2,3}} \textnormal{and} 
         Sophia Ananiadou\textsuperscript{\textnormal{1,3,4}}\\  
         \textsuperscript{\textnormal{1}}National Centre for Text Mining, Department of Computer Science, \\ The University of Manchester, United Kingdom \\
         \textsuperscript{\textnormal{2}}Toyota Technological Institute, Japan \\
         \textsuperscript{\textnormal{3}}Artificial Intelligence Research Center (AIRC),\\
  National Institute of Advanced Industrial Science and Technology (AIST), Japan \\
  \textsuperscript{\textnormal{4}}Alan Turing Institute, United Kingdom \\
  {\tt \{nhung.nguyen, sophia.ananiadou\}@manchester.ac.uk} \\
  {\tt makoto-miwa@toyota-ti.ac.jp }
 }
 
\begin{document}
\setlength{\abovedisplayskip}{6pt}
\setlength{\belowdisplayskip}{6pt}
\setlength{\abovedisplayshortskip}{6pt}
\setlength{\belowdisplayshortskip}{6pt}
\maketitle
\begin{abstract}
The information bottleneck (IB) principle has been proven effective in various NLP applications.
The existing work, however, only used either generative or information compression models to improve the performance of the target task. 
In this paper, we propose to combine the two types of IB models into one system to enhance Named Entity Recognition (NER).
For one type of IB model, we incorporate two unsupervised generative components, span reconstruction and synonym generation, into a span-based NER system.
The span reconstruction ensures that the contextualised span representation keeps the span information, while the synonym generation makes synonyms have similar representations even in different contexts. 
For the other type of IB model, we add a supervised IB layer that performs information compression into the system to preserve useful features for NER in the resulting span representations.
Experiments on five different corpora indicate that jointly training both generative and information compression models can enhance the performance of the baseline span-based NER system.
Our source code is publicly available at \url{https://github.com/nguyennth/joint-ib-models}.
\end{abstract}

\section{Introduction}
\citet{tishby99information} introduced the information bottleneck (IB) method to compress representation while preserving meaningful information.
The method has been incorporated in many state-of-the-art (SOTA) deep models such as Variational Autoencoder (VAE)~\cite{Kingma2014,pmlr-v32-rezende14}, 
and Deep Variational Information Bottleneck (Deep VIB)~\cite{alemi2016variational}. 
Those deep models can be divided into supervised generative models (e.g., Deep VIB) and unsupervised ones (e.g., VAEs)~\cite{vib}.

Both VAE and VIB have been applied to NLP applications.
For example, \citet{CRF-VAEs} and \citet{chen-etal-2018-variational} proposed to use VAE in sequence labelling tasks such as POS tagging and NER.
Meanwhile, \citet{wang-etal-2022-miner} used VIB to tackle OOV issues in NER.
The two types of IB have also been employed in other tasks, such as dialogue response generation~\cite{chen-etal-2022-dialogved}, parsing~\cite{li-eisner-2019-specializing}, paraphrase generation for MT~\cite{ormazabal-etal-2022-principled}, and text summarisation~\cite{west-etal-2019-bottlesum}, to name a few.
Such previous work only used one type of IB model (either VAE or VIB) in their system, and it is unclear whether we can effectively combine the two types of IB models.

The NER task has been typically approached using sequence models such as BiLSTM~\cite{lample-etal-2016-neural} and BERT~\cite{devlin-etal-2019-bert,10.1093/bioinformatics/btz682,beltagy-etal-2019-scibert}.
At the same time, we have seen the rise of span-based models~\cite{sohrab-miwa-2018-deep,zheng-etal-2019-boundary,Tan_Qiu_Chen_Wang_Huang_2020,xia-etal-2019-multi,Xu_Huang_Feng_Hu_2021,fu-etal-2021-spanner,li-etal-2021-span}, which are simple and effective. 
Using a span-based model, we can directly represent and manipulate span representations.

This paper investigates the effects of combining the two IB types for the NER task.
To that end, we jointly train the span-based NER system with two VAE components and one VIB component. 
The first VAE component is span reconstruction, used to reconstruct original spans.
This component is similar to Sentence VAEs~\cite{bowman-etal-2016-generating}, but the model learns only from spans instead of sentences.
The second VAE one is synonym generation, used to generate synonym(s) of a span. 
We first collect synonyms of each span from an external knowledge base (KB) and then train a VAE model that can generate the corresponding synonyms given a span.
By adding the synonym generation into the model, we indirectly inject semantic information from synonyms into the span representation.
The last component is a VIB component~\cite{mahabadi2021variational} introduced into the system to compress span representations while keeping useful features for NER.

We evaluate the proposed model on five different corpora: BC5CDR~\cite{BC5CDR}, GENIA~\cite{GENIA}, MedMention-21st~\cite{MedMention}, NCBI Disease~\cite{NCBI_Disease}, and ShARe/CLEFE~\cite{ShareClef}.
For the synonym generation component, we identify synonyms of each span by performing exact matching against mentions in Unified Medical Language System (UMLS)~\cite{Bodenreider04}.
Experimental results show that by incorporating the two IB types into a span-based NER, we can improve the performance over the baseline span-based model on BC5CDR-Disease, GENIA, MedMention, and NCBI datasets.
In the case of GENIA--one of the most popular corpora for nested entities, the proposed model could perform favourably compared with current SOTA systems, even with the simple span-based baseline model without any recent enhancements like boundary detection~\citep{Tan_Qiu_Chen_Wang_Huang_2020,Xu_Huang_Feng_Hu_2021}.
Furthermore, such boundary-enhanced models can benefit from our approach. 

We additionally performed some analysis on the intermediate output of the proposed model.
We observed that when jointly trained VAEs with a NER task, the latent variable is restructured satisfactorily towards the task.
Similarly, posteriors estimated by the VIB component are clustered neatly even though the input information is compressed. 
Such distinguishable clusters are potentially helpful for entity linking~\cite{liu-etal-2021-self}.
We also found that synonym generation helped improve the quality of span reconstruction and the NER performance.

In summary, the contributions of our paper are:
\setlist{nolistsep}
\begin{itemize}[noitemsep]
    \item This is the first study that investigates the impact of combining two IB methods on NER. 
    \item Through experiments on five different corpora, we demonstrate that the joint model can improve the baseline performance in most cases.
    \item In-depth analyses on the intermediate output of the joint model indicate that each component plays a different role in enhancing span representations as expected. 
\end{itemize}

\section{Related Work}
\subsection{Span-based NER}
Traditionally, sequence models such as BiLSTM~\cite{lample-etal-2016-neural} and BERT~\cite{devlin-etal-2019-bert,10.1093/bioinformatics/btz682,beltagy-etal-2019-scibert} have been used to tackle the task of NER, producing state-of-the-art (SOTA) performance.
However, those models could not perform on overlapping entities, i.e., a span has more than one named entity category or nested entities.
To address this issue, \citet{sohrab-miwa-2018-deep} proposed the span-based approach.
In this approach, all possible spans are exhaustively generated given a specific span length. 
The span representation was calculated based on a pre-trained language model and then classified to a corresponding entity type by a linear layer.
Following the suite, numerous studies have shown that span-based approaches to NER could produce SOTA performance~\cite{zheng-etal-2019-boundary,Tan_Qiu_Chen_Wang_Huang_2020,xia-etal-2019-multi,sohrab-etal-2019-neural,Xu_Huang_Feng_Hu_2021,fu-etal-2021-spanner,li-etal-2021-span,Yu2022}.

\citet{fu-etal-2021-spanner} designed SpanNER that learns the representation of a span based on its token representation and the span length embedding. SpanNER can also make ensemble predictions from both span-based and sequence labelling systems. 
Similarly, \citet{Yu2022} proposed SNER that represents a span by considering the context embedding from BERT, i.e., the CLS embedding. 
\citet{ouchi-etal-2020-instance} proposed to learn the similarity between spans using instance-based learning. The model will assign the class label based on its similar training span at inference time.
\citet{Xu_Huang_Feng_Hu_2021} used a supervised multi-head self-attention mechanism, where each head corresponds to one entity type, to construct the word-level correlations for each type. 

Following the work mentioned above, this paper also focuses on span-based models.
Our model, however, learns span representation differently.
In particular, we update the span representation using span reconstruction and synonym generation.
We then compress it by selecting relevant features using the IB method. 
It is noted that our model can potentially be incorporated into existing span-based models.

\begin{figure*}
    \centering
    \includegraphics[width=0.825\textwidth]{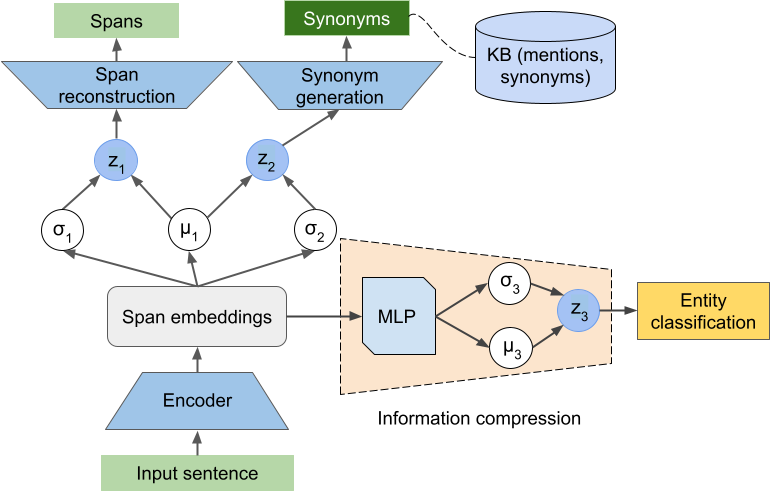}
    \caption{Span-based NER is jointly trained with one supervised VIB component: information compression, and two unsupervised VAE components: span reconstruction and synonym generation. The encoder is shared across the components. Synonyms are collected from an external knowledge base (KB).}
    \label{fig:model}
\end{figure*}

\subsection{Information Bottleneck in NLP}
The Information Bottleneck (IB) method has been applied to various NLP applications.
We divide those studies into two main groups: generative models and information compression models.

In the first group, we have studies that used IB as generative models, namely Variational Autoencoders (VAEs).
\citet{CRF-VAEs} proposed a sequence labelling NER model that treats a neural CRF as the amortised approximate posterior in a discrete structured VAE.
Meanwhile, \citet{chen-etal-2018-variational} applied neural variational methods to sequence labelling by combining a latent variable generative model and a discriminative labeller. 
VAEs were also employed in dialogue response generation~\cite{chen-etal-2022-dialogved} and relation extraction~\cite{yuan-eldardiry-2021-unsupervised,christopoulou-etal-2021-distantly}.

The second group are studies that used IB as information compression, namely Variational Information Bottleneck (VIB). 
\citet{mahabadi2021variational} was one of the first studies using VIB to fine-tune low-resource target tasks.
\citet{wang-etal-2022-miner} employed the information bottleneck principle to tackle OOV issues in NER.
Information compression was also used in parsing~\cite{li-eisner-2019-specializing}, paraphrase generation for MT~\cite{ormazabal-etal-2022-principled}, research replication prediction~\cite{luo-etal-2022-interpretable}, text classification~\cite{zhang-etal-2022-improving}, improving the attention's reliability~\cite{zhou-etal-2021-attending} and text summarisation~\cite{west-etal-2019-bottlesum}.

Unlike previous studies, we combine generative and information compression models for the task of NER. 
In our model, span reconstruction ensures that the span representation keeps the span information, while synonym generation makes similar synonyms have similar representations even in different contexts. 
Meanwhile, information compression helps suppress irrelevant features, addressing overfitting (if any). 
To the best of our knowledge, this is the first paper investigating such a joint system for span-based NER.

\section{Methods}
The overall framework of our model is illustrated in Figure~\ref{fig:model}.
We first encode span embeddings using a transformer-based network.
The output from the encoder is then used in three components: (1) entity classification that classifies input spans into an entity type or non-entity using a supervised VIB framework~\cite{mahabadi2021variational}; (2) span reconstruction that recovers input spans of gold entities; and (3) synonym generation that generates synonyms for a gold entity.
Similarly to a multi-task setting, we simultaneously train the three components.
We only run the first component at the inference step to predict named entities.

\subsection{Span Reconstruction and Synonym Generation}
\subsubsection{Encoder}
Given an input sentence with $n$ words: $\{w_0, w_1, ..., w_n\}$, we use a multi-layer transformer-based encoder~\cite{devlin-etal-2019-bert} to encode the sentence. 
As a result, we have the following contextualised sub-word vectors $\{\bm{v}_0, \bm{v}_1, ..., \bm{v}_m\}$.
Similarly to \newcite{sohrab-miwa-2018-deep}, we then exhaustively extract all possible spans with a maximum length of $sl$ from the input sentence. 
Each span embedding is calculated as follows:
\begin{equation}
    \bm{s}_{i,j}= \left[\bm{v}_i;\frac{\Sigma_{t=i}^{j}\bm{v}_t}{j-i+1};\bm{v}_j\right],
\label{eq:span}
\end{equation}
where $i$ and $j$ are the start and end positions of the span, $\bm{v}_t$ is the embedding of the $t$-th sub-word, and $[;;]$ denotes the concatenation operation.

We then apply two linear layers on top of the span embeddings to construct the parameters of a posterior distribution $q(\bm{z}|\bm{s})$ using the following equations:
\begin{equation}
\begin{aligned}
    \bm{\mu} = \bm{W}_{\mu}\bm{s} + \bm{b}_{\mu}, \\
    \bm{\sigma}^2 = \bm{W}_{\sigma}\bm{s} + \bm{b}_{\sigma}, 
\end{aligned}
\end{equation}
where $\bm{\mu}$ and $\bm{\sigma}$ are the parameters of a multivariate Gaussian, representing the feature space of the span; $\bm{W}$ and $\bm{b}$ are weights and biases of the linear layers, respectively. 
The posterior distribution is approximated via a latent variable $\bm{z}$ using the reparameterisation trick~\cite{Kingma2014} as follows
\begin{equation}
    \bm{z} = \bm{\mu} + \bm{\sigma}\bm{\epsilon}, \text{where } \bm{\epsilon} \sim \bm{\mathcal{N}}(0,1).
    \label{eq:repara}
\end{equation}

In the case that we train both span reconstruction and synonym generation, we have different parameters of $\bm{\sigma}_1$ and $\bm{\sigma}_2$, respectively. 
To encourage span and synonyms to distribute closely, $\bm{\mu}$ is shared between them.

\subsubsection{Decoders}
Our decoder is an LSTM network that greedily reconstructs the input span or generates all the corresponding synonyms (if any) in an autoregressive manner.
Given latent $z$ from the encoder, we first use $\bm{z}$ to initialise the hidden state of the decoder via a linear layer transformation.
We then form the input of the decoder with the teacher forcing strategy~\cite{williams1989learning}, i.e., we concatenate $\bm{z}$ with the representation of each word $w_t$\footnote{We pass a special start symbol as $w_0$.} in a given gold span (for span reconstruction) or a gold synonym (for synonym generation).

\subsubsection{Learning}
To train VAEs, we maximise the Evidence Lower BOund (ELBO) that includes two losses: the reconstruction loss and the Kullback-Leibler divergence ($D_{KL}$).
The reconstruction loss is the cross entropy loss between an actual span and its reconstruction or synonym.
The $D_{KL}$ is calculated based on a prior distribution ($p(\bm{z})$) and the posterior distribution ($q(\bm{z}|\bm{s})$) produced by the decoder.
In the case of span reconstruction (SR), the loss for a span $s$ is 
\begin{equation}
\begin{aligned}
    L_{SR}(\bm{\theta}_1,\bm{\phi}_1) = \underset{\bm{z}_1 \sim q_{\bm{\phi}_1}(\bm{z}_1|\bm{s})}{\mathbb{E}}[\log(p_{\bm{\theta}_1}(\bm{s}|\bm{z}_1))] \\
    - D_{KL}(q_{\bm{\phi}_1}(\bm{z}_1|s)\parallel p_{\bm{\theta}_1}(\bm{z}_1)).
\end{aligned}
\label{eq:spanreco}
\end{equation}
Similarly, with the synonym generation (SG), the loss is
\begin{equation}
\begin{aligned}
    L_{SG}(\bm{\theta}_2,\bm{\phi}_2) = \underset{\bm{z}_2 \sim q_{\bm{\phi}_2}(\bm{z}_2|\bm{s})}{\mathbb{E}}[\log(p_{\bm{\theta}_2}(\bm{s}|\bm{z}_2))] \\
    - D_{KL}(q_{\bm{\phi}_2}(\bm{z}_2|\bm{s})\parallel p_{\bm{\theta}_2}(\bm{z}_2)).
\end{aligned}
\label{eq:syngen}
\end{equation}
In Equations~\ref{eq:spanreco} and \ref{eq:syngen}, $\bm{\theta}$ and $\bm{\phi}$ are weights and biases of the network, respectively.
We attach a subscript to each parameter to denote that the parameter belongs to the span reconstruction component or the synonym generation one.

\subsection{Entity Classification with Supervised IB}
The main objective of the supervised IB is to preserve the information about the target class(es) in the latent while filtering out irrelevant information from the input~\cite{vib}.
As a result, the objective loss function for supervised IB is based on the compression loss and the prediction loss, as shown in Equation~\ref{eq:vib}.
\begin{equation}
\begin{aligned}
    L_{VIB}(\bm{\theta}_3,\bm{\phi}_3) &= \beta \text{ } \underset{s}{\mathbb{E}}[D_{KL}(p_{\bm{\theta}_3}(\bm{z}_3|\bm{s}),r(\bm{z}_3))] \\
    & + \underset{\bm{z}_3 \sim p_{\bm{\theta}_3}(\bm{z}_3|\bm{s})}{\mathbb{E}}[-\log{q_{\bm{\phi}_3}}(y|\bm{z}_3)],
\label{eq:vib}
\end{aligned}
\end{equation}
where $r(\bm{z}_3)$ is an estimate of the prior probability $p_{\bm{\theta}_3}(\bm{z}_3)$, $\beta$ is in a range of $[0,1]$, and $y$ is the true label of the input span.
Similarly to previous equations, $\bm{\theta}_3$ and $\bm{\phi}_3$ are weights and biases of the network, respectively.
\begin{table*}[t]
    \centering
    \small
    \begin{tabular}{|l|rrr|rrr|rrr|l|}
    \hline
        \multirow{2}{*}{Corpus} &  
        \multicolumn{3}{c|}{Train} & 
        \multicolumn{3}{c|}{Dev} & 
        \multicolumn{3}{c|}{Test} & 
        \multirow{2}{*}{Entity}\\ \cline{2-10}
        \multicolumn{1}{|c|}{} & \#Doc. & \#Ent. &Syn\_cov & \#Doc. & \#Ent. &Syn\_cov & \#Doc. & \#Ent. &Syn\_cov & \multicolumn{1}{|c|}{} \\ \hline
        NCBI & 592 & 5,134 & 79.25\% & 100 & 787 & 74.46\%& 100 & 960 & 75.62\%& Disease \\ \hline
        BC5-Dis & 500 & 4,182 & 87.18\% & 500 & 4,244 & 87.18\% & 500 & 4,424 & 89.69\% & Disease \\ \hline
        BC5-Chem & 500 & 5,203 & 93.14\% & 500 & 5,347 & 94.58\% & 500 & 5,385 & 94.00\% & Chemical \\ \hline
        ShARe & 149 & 3,630 & 77.96\% & 50 & 1,413 & 77.35\% & 99 & 4,912 & 81.54\% & Disorder \\\hline
        GENIA & 1,599 & 45,036 & 55.98\% & 189 & 4,274 & 49.46\% & 212 & 5,346 & 51.05\% & 5 types  \\\hline
        MM & 2,635 & 122,241 & 70.21\% & 878 & 40,884 & 70.55\% & 879 & 40,157 & 71.17\%& 21 types \\ \hline
    \end{tabular}
    \caption{Statistics numbers of experimental corpora. \#Doc. indicates the number of documents; \#Ent. indicates the number of gold entities; Syn\_cov indicates the percentage of gold entities that have synonyms in UMLS.}
    \label{tab:data}
\end{table*}

Following \newcite{mahabadi2021variational}, we use a multi-layer perceptron (MLP) with two linear layers to compute the compressed representation of a span.
We approximate $q_{\bm{\phi}_3}(y|\bm{z}_3)$ using another linear layer.
In particular, this layer is a binary classifier with a sigmoid function to predict the correct entity category of an input span. 
We use binary cross entropy (BCE) loss to compute the prediction loss (i.e., the second term of Equation~\ref{eq:vib}) in our model.

\subsection{Training Objective}
We jointly train span reconstruction, synonym generation, and entity classification with the sum of the following optimisation objective for all the spans:
\begin{equation}
    L = L_{VIB} + \gamma(L_{SR} + L_{SG}),
\end{equation}
where $\gamma$ is a hyper-parameter with a range of $[0,1]$. 
It is noted that during training, $L_{VIB}$ is calculated for all spans, while $L_{SR}$ and $L_{SG}$ are calculated for gold spans only.

\section{Experiments}
\subsection{Datasets}
We conducted experiments on five different datasets: NCBI Disease~\cite{NCBI_Disease}, BC5CDR~\cite{BC5CDR}, ShARe/CLEFE (ShARe)~\cite{ShareClef}, GENIA~\cite{GENIA}, and MedMention-21st (MM)~\cite{MedMention}.
Following previous work, we divided BC5CDR into BC5CDR-Disease (BC5-Dis) and BC5CDR-Chemical (BC5-Chem).
To detect synonyms of each span, we only performed exact matching against mentions from UMLS~\cite{Bodenreider04}. 
In Table~\ref{tab:data}, we report the number of documents, the number of golden entities, and the percentage of golden entities that have synonyms in the UMLS (version 2017AA) of the experimental corpora. 
Since UMLS does not cover every entity, there are cases that an entity does not have any synonyms\footnote{In this case, we only lowercase texts in the data and UMLS before matching. Using other techniques to find synonyms in UMLS is beyond the scope of this paper.}.

\subsection{Settings}
Regarding the encoder, we employed the pre-trained SciBERT model~\cite{beltagy-etal-2019-scibert}. 
Regarding the decoder, we used an LSTM with one hidden layer; input vectors to the LSTM were extracted from resulting vectors trained on a combination of PubMed, PMC texts and the English Wikipedia~\cite{Pyysalo2013DistributionalSR}.

It is noted that before jointly training with the entity classification, we pre-trained the auxiliary components, i.e., span reconstruction and synonym generation in several epochs. 
The NER performance was measured based on the exact matching F1 scores calculated by the N2C2 Shared Task NER evaluation script~\cite{n2c2-2018}.
All hyper-parameter settings are detailed in Appendix~\ref{apd:para}.

\subsection{Results}
Across all corpora, we compare the joint model in Figure~\ref{fig:model} with a Baseline system that is a span-based model\footnote{We used the span-based NER implementation by \citet{trieu:2020}} using SciBERT, as reported in Table~\ref{tab:ner_results}.
With the GENIA corpus, we also collected the performance of other span-based SOTA systems for comparison, as shown in Table~\ref{tab:genia}.
\begin{table}[t]
    \centering
    \begin{tabular}{|l|c|c|}
        \hline
         Corpus & Baseline & Joint model \\
         \hline
         BC5-Chemical &91.38 &91.30 \\
         BC5-Disease &  84.46 &\textbf{85.25} \\
         GENIA & 76.93 & \textbf{77.43} \\
         MedMention-21st & 62.78 & \textbf{63.21} \\
         NCBI  &88.12 & \textbf{88.29}\\
         ShARe/CLEFE & 80.88 & 80.77 \\
         \hline
         Average & 80.76 & \textbf{81.04} \\
         \hline
    \end{tabular}
    \caption{F1 scores (\%) on the test sets. Bold numbers indicate the joint model is better than the baseline. 
    }
    \label{tab:ner_results}
\end{table}

\begin{table}[]
    \centering
    \begin{tabular}{|l|c|}
    \hline
        System & F1 score \\
    \hline
    Instance-based \cite{ouchi-etal-2020-instance} & 74.20 \\
    Boundary-aware \cite{zheng-etal-2019-boundary} & 74.70 \\
    BENSC \cite{Tan_Qiu_Chen_Wang_Huang_2020} & 78.30\\
    MHSA \cite{Xu_Huang_Feng_Hu_2021} & 79.60 \\
     \hline\hline
    Baseline & 76.93 \\
    Joint model & \textbf{77.43} \\
    \hline
    \end{tabular}
    \caption{F1 scores (\%) on the GENIA testing set compared with other SOTA systems. 
    }
    \label{tab:genia}
\end{table}

In Table~\ref{tab:ner_results}, we can see that performance across the corpora differs. 
The proposed model produced better NER performance than the baseline ones on four corpora.
The exception happened with BC5CDR-Chemical and ShARe/CLEFE where the simple span-based model obtained higher F1 scores than the joint model.
The corpora's characteristics are one possible reason for this. For BC5-Chemical, reconstructing chemical entities and finding their correct synonyms are more complex than the other entities (see Section~\ref{sec:generation}). For ShAe/CLEFE, its documents are health records, not scientific papers like the others. 
On average, however, the joint model performed better than the baseline.

Results in Table~\ref{tab:genia} show that on the GENIA corpus, the joint model could perform better than (1) the baseline and (2) the two span-based models: the instanced-based NER model~\cite{ouchi-etal-2021-instance} and the boundary-aware one~\cite{zheng-etal-2019-boundary}. However, our model produced lower F1 scores than BENSC~\cite{Tan_Qiu_Chen_Wang_Huang_2020} and MHSA~\cite{Xu_Huang_Feng_Hu_2021}.
This can be explained by the fact that both BENSC and MHSA specifically address nested entities by enhancing span boundary detection, while our model classifies all possible spans.
Nevertheless, it is noted that our objective is not to improve the SOTA; we focus on investigating the joint model for the NER task anyway.
Furthermore, our model can be incorporated into those span-based SOTA systems.

\section{Analysis}
\subsection{Ablation Study}
To evaluate the effect of each component on the proposed model, we ran the following settings on the development sets:
\setlist{nolistsep}
\begin{itemize}[noitemsep]
    \item SupVIB: the span-based model using the supervised IB entity classification.
    \item SupVIB\_0: $\beta$ was set to 0, meaning that the compression loss was not involved in the training stage. This setting is similar to the baseline setting, but the span embeddings are compressed by the MLP before being fed into the linear layer.
    \item SupVIB + SpanReco: we jointly trained span reconstruction with the SupVIB.
    \item SupVIB + SpanReco + SynGen (All): the full joint model with all three components.
\end{itemize}

From Table~\ref{tab:dev_ablation}, we can observe the following.
Firstly, using SupVIB, we could obtain higher F1 scores in most of the corpora than the baseline.
The compression loss is important for SupVIB. In four over six corpora, setting $\beta = 0$ made the performance drop.
\begin{table*}[t]
    \centering
    \begin{tabular}{|l|c|c|c|c|c|c|c|}
    \hline
        Setting &  BC5-Chem & BC5-Dis & GENIA & MM & NCBI & ShARe & Avg.\\
    \hline
        Baseline & 93.34 & 84.88 & 78.00 & 62.83 & 87.89& 80.89 & 81.30\\
        SupVIB\_0 & 94.01 & 85.51& 78.26& 62.31& 87.76 & 81.14 & 81.50\\
        SupVIB & \textbf{94.05} & 85.51 & 78.36 & 63.00 & 87.70 & 81.65 & 81.71\\
        SupVIB + SpanReco & 93.69 & 85.23 & 78.22 & \textbf{63.63} & 88.05 & 81.56 & 81.73 \\
        All & 93.78 & \textbf{85.56} & \textbf{79.30} & 63.39 & \textbf{88.41} & 81.11 & \textbf{81.93}\\
    \hline
    \hline
    All (w/ shared $\bm{\sigma})$ & 93.87 & 85.33 & 78.90 &63.21 & 87.62 &80.93 & 81.64 \\
    All (w/o shared $\bm{\mu}$ and $\bm{\sigma})$ & 93.49 & 85.03 & 78.97 & 61.85 & 87.89 & \textbf{81.78} & 81.50 \\
    \hline
    \end{tabular}
    \caption{F1 scores (\%) on the development sets produced by different settings. All is the Joint model in Tables~\ref{tab:ner_results} and \ref{tab:genia}. The last column is the average score across the corpora. Bold numbers indicate the best performance in each corpus.}
    \label{tab:dev_ablation}
\end{table*}
Secondly, when we introduced the span reconstruction component alone into the model, except for MedMention-21st and NCBI, the component degraded the performance. 
We observed that SupVIB restructured embeddings differently than SupVIB+SpanReco, which probably affected the performance. Moreover, SupVIB was slightly better than SupVIB+SpanReco in distinguishing entities' boundaries. 

When we introduced synonym generation into the model (i.e., the All setting), we could obtain better performance than the baseline model on five over six corpora. This indicates that indirectly injecting semantic information into the span representation is mostly helpful. The exceptions were with MedMention and ShARe/CLEFE, where the All setting produced a lower F1 score than the SupVIB+SpanReco.

By jointly training two types of IB, i.e., generative model and supervised IB, we could enhance the NER performance over the baseline on four out of six testing corpora. Especially with the two most complex, multi-category NER corpora, i.e., GENIA and MedMention-21, the joint model could outperform the baseline by a larger margin than the other corpora.

Lastly, looking at the last column in Table~\ref{tab:dev_ablation}, i.e., the average F1 score across the corpora, we can see that the scores were increased from the top to the bottom. We, therefore, can confirm that adding the proposed components one after another helped improve the performance in general. 

\subsubsection{Impact of the Shared Parameters in VAEs}
To encourage spans and their synonyms to distribute closely, we have $\bm{\mu}$ shared between the span reconstruction and the synonym generation, as in the aforementioned All model.
Theoretically, we can also share $\bm{\sigma}$ between the two components.
To investigate the impact of these shared parameters, we conducted two more experiments: 
\begin{itemize}
    \item All with shared $\bm{\sigma}$: the full model in which the two VAE components share $\bm{\mu}$ and $\bm{\sigma}$.
    \item All without shared $\bm{\mu}$ and $\bm{\sigma}$: the full model in which each VAE component has its own (i.e., independent) $\bm{\mu}$ and $\bm{\sigma}$.
\end{itemize}

From the last two rows in Table~\ref{tab:dev_ablation}, we can see that the All model with shared $\bm{\sigma}$ made the F1 scores drop across the corpora (except for the BC5CDR-Chemical) compared with the All model.
This is expected because sharing $\bm{\mu}$, i.e., sharing the mean of the two distributions, is more reasonable than sharing both of the parameters, i.e., sharing both the mean and the variance of the two distributions.
However, when we have the independent $\bm{\mu}$ and $\bm{\sigma}$, there is no common pattern in the performance.
Compared with the All model, the All without shared $\bm{\mu}$ and $\bm{\sigma}$ model produced lower scores on most of the corpora, except for the ShARe/CLEFE one.
On average, the shared $\bm{\mu}$ setting (i.e., the All model) was superior to the other two.

\subsection{NER Error Analysis}
We classified false positive predictions by NER models into two classes:
\begin{itemize}
    \item Category errors: denote predictions that have correct spans but wrong category,
    \item Span errors: denote predictions that have incorrect spans.
\end{itemize}

Regarding the BC5CDR, NCBI, and ShARe/CLEFE corpora, all false positives are due to span errors since these corpora have only one NER category. 
We report the percentage of span errors in false positives on the development sets of these corpora in Table~\ref{tab:FP-spanerror}.
On BC5-Chemical and ShARe/CLEFE, the proposed model produced more span errors than the baseline one.
This situation somehow aligns with the exceptionally lower performance of the proposed model on the two corpora reported in Table~\ref{tab:ner_results}.
\begin{table}[t]
    \centering
    \begin{tabular}{|l|r|r|}
    \hline
        Corpus &  Baseline & \multicolumn{1}{c|}{All}\\
    \hline
        BC5-Chemical & \bf{6.48} & 6.50 \\
        BC5-Disease & 15.21 & \bf{13.52} \\
        NCBI & 12.76 & \bf{12.36} \\
        ShARe/CLEFE & \bf{18.02} & 19.23 \\
    \hline
    \end{tabular}
    \caption{Percentage of span errors on the development sets. The lower the number, the better the model.}
    \label{tab:FP-spanerror}
\end{table}

In cases of MedMention-21st and GENIA, we report the number of category and span errors in Table~\ref{tab:FP-all}.
We can see that the proposed model produced fewer category errors than the baseline one on both corpora.
We hypothesise that the synonym generation component attributed to this improvement; the semantic information provided by the component enhanced the ability to distinguish named entities' categories of the All model. 
Moreover, we find that categories in GENIA are more ambiguous than those in MedMention-21st.
For example, it is difficult to distinguish between ``Cell line'' and ``Cell type'' or between ``Protein'' and ``DNA''. 
This can explain the lower reduction of category errors in GENIA than in MedMention-21st.

\begin{table}[t]
    \centering    
    \begin{tabular}{|l|r|r|r|r|}
    \hline
    \multirow{2}{*}{Corpus} & \multicolumn{2}{c|}{Category Error} & \multicolumn{2}{c|}{Span Error}\\
    \cline{2-5} 
        & Base & All & Base & All  \\
    \hline
       GENIA & 126 & \bf{115} & \bf{4,133} &	4,411  \\
       MM & 2,447 & \bf{2,028} & 37,148  & \bf{34,747} \\
    \hline
    \end{tabular}
    \caption{Number of category errors and span errors on the development sets. Base means the Baseline model. The lower the number, the better the model.}
    \label{tab:FP-all}
\end{table}

Table~\ref{tab:FP-all} also shows that the proposed model could help reduce span errors on MedMention-21st but not on GENIA. 
We think that the effectiveness of our model depends on the annotation schemes. 
As mentioned above, the GENIA corpus contains gene/protein-related entities, of which the boundary of an entity is more specially defined than those in MedMention-21st.
As a result, detecting correct spans in GENIA is more challenging than in MedMention. 
For example, the All model detected ``HBxAg - specific synthetic polypeptides'' as a Protein entity while the correct span should be ``HBxAg'' only.

Nevertheless, the span errors accounted for a majority of the false positives across all corpora, indicating that an enhanced-boundary approach can help alleviate the situation.

\subsection{Generation Quality}
\label{sec:generation}
To evaluate the generation quality of the VAEs component, we calculated BLEU-2 scores of the reconstructed gold entities on the development set of each corpus.
From Table~\ref{tab:bleu2}, we can see that using synonym generation helped improve the BLEU scores in almost corpora.
This indicates that semantic information from the synonym generation component is useful for the NER task and the span reconstruction. 
\begin{table}[t]
    \centering
    \begin{tabular}{|l|c|c|}
    \hline
        Corpus &  \multicolumn{1}{p{1.7cm}|}{SupVIB+ SpanReco} & All \\
    \hline
        BC5-Chemical & 0.0066 & 0.0051 \\
        BC5-Disease & 0.0092 & \textbf{0.0243} \\
        GENIA & 0.3902 & \textbf{0.4479} \\
        MedMention-21st & 0.0612 & \textbf{0.1482} \\
        NCBI & 0.1244 & \textbf{0.2228} \\
        ShARe/CLEFE & 0.0033 & \textbf{0.1283} \\      
    \hline
    \end{tabular}
    \caption{BLEU-2 scores of span reconstruction on the
development set of each corpus.}
    \label{tab:bleu2}
\end{table}
\begin{table*}[t]
    \centering
    \begin{tabular}{|r|p{4.2cm}|p{4cm}|p{4cm}|}
    \hline
    No. & Original span & SupVIB + SpanReco & All \\
    \hline
    1 & nervous fibers & central channel & nervous muscles \\
    2 & renal failure & pain failure & renal failure \\
    3 & immunomodulatory therapy & Cg therapy & antiviral therapy \\
    4 & central hydrophobic core & central its carbon & central temporal core \\
    5 & pulmonary veins	& liver ganglion & pulmonary vein \\
    6 & fibre degeneration & hearing loss & macular degeneration \\
    7 & cardiac inter - beat & cardiac - time coil & cardiac - - range  \\
    8 &  volatile compounds	& translational compounds & volatile compounds \\
    9 & PPA & antidepressant & PPA \\
    10 &  autoimmune lymphoproliferative syndrome & ankylosing lymphoproliferative syndrome & autoimmune lymphoproliferative syndrome \\
    \hline
    \end{tabular}
    \caption{Some examples to show that the synonym information is helpful to span reconstruction.}
    \label{tab:reco}
\end{table*}

In Table~\ref{tab:reco}, we show examples where having synonym generation, i.e., the All model, could reconstruct original entities much better than SupVIB+SpanReco.
In all examples, the reconstructed spans by the All model are more meaningful and fluent than those by SupVIB+SpanReco.

Among the six corpora, reconstructing chemical entities is the most challenging task, especially with short-form chemicals such as PPA (Phenylpropanolamine).
When looking for synonyms of these entities in UMLS, we only did exact matching without checking any semantics.
Therefore, there are cases in which we found UMLS synonyms for these abbreviations, but their full forms are completely different.
For example, some UMLS synonyms of PPA are ``primary progressive aphasia'', ``primary progressive apraxia of speech'', and ``Mesulam syndrome'', which are not relevant to ``Phenylpropanolamine''---the correct full form of the chemical PPA.
Nevertheless, we observed that the All model usually reconstructed an abbreviation if the original entity is an abbreviation, while the SupVIB+SpanReco did it very randomly. 
For example, in row 9 in Table~\ref{tab:reco}, given an abbreviation of ``PPA'', the All model could generate the exact full form given span, while SupVIB+SpanReco generated ``antidepressant''--an incorrect full form for ``PPA''.

\begin{figure*}
    \centering
    \begin{subfigure}[b]{0.22\textwidth}
        \centering
        \includegraphics[width=\textwidth]{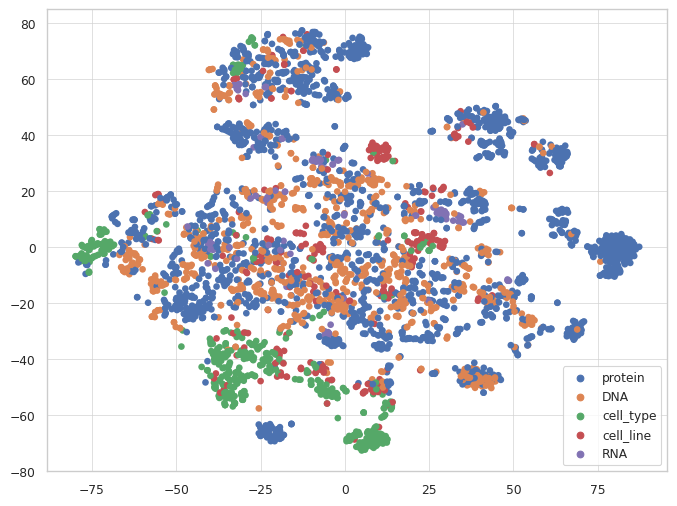}
        \caption{GENIA: Only VAEs}
        \label{fig:genia-vae}
    \end{subfigure}
    \begin{subfigure}[b]{0.22\textwidth}
        \centering
        \includegraphics[width=\textwidth]{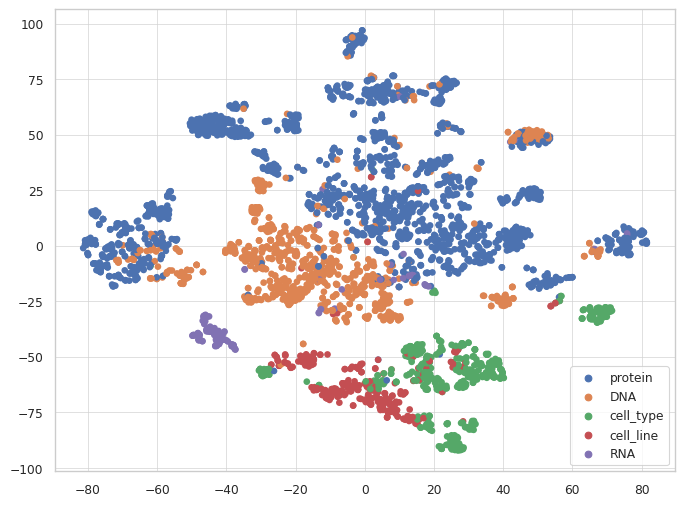}
        \caption{GENIA: The joint model}
        \label{fig:genia-joint}
    \end{subfigure}
    \begin{subfigure}[b]{0.22\textwidth}
         \centering
         \includegraphics[width=\textwidth]{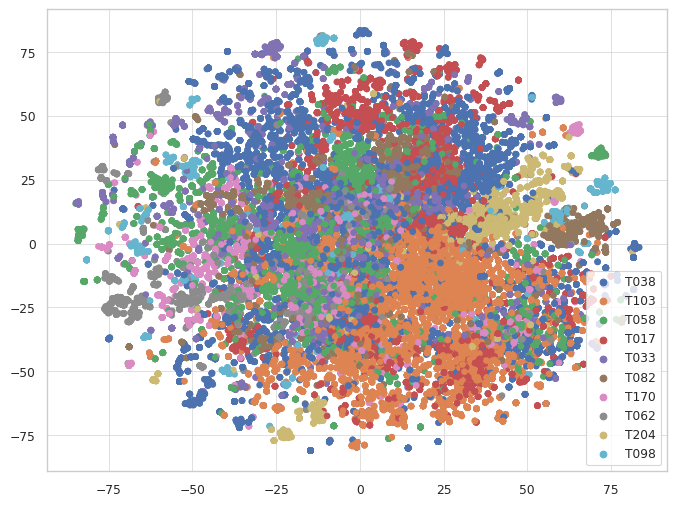}
         \caption{MM21: Only VAEs}
         \label{fig:mm21-vae}
     \end{subfigure}
    \begin{subfigure}[b]{0.22\textwidth}
         \centering
         \includegraphics[width=\textwidth]{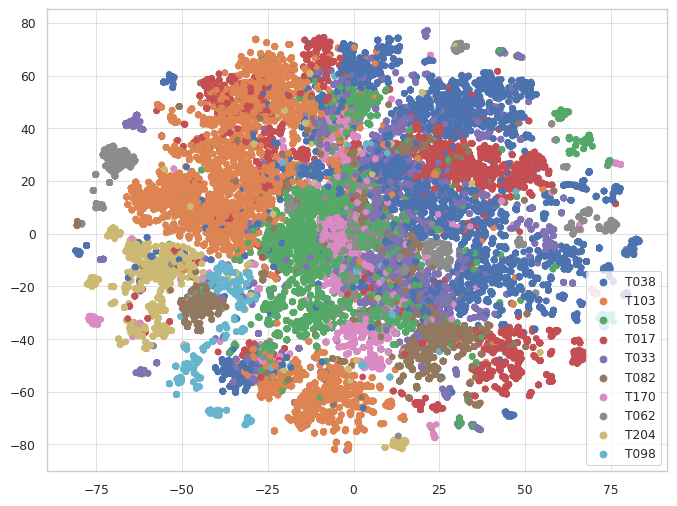}
         \caption{MM21: The joint model}
         \label{fig:mm21-joint}
     \end{subfigure}
     
    \bigskip 
    \centering
    \begin{subfigure}[b]{0.22\textwidth}
        \centering
        \includegraphics[width=\textwidth]{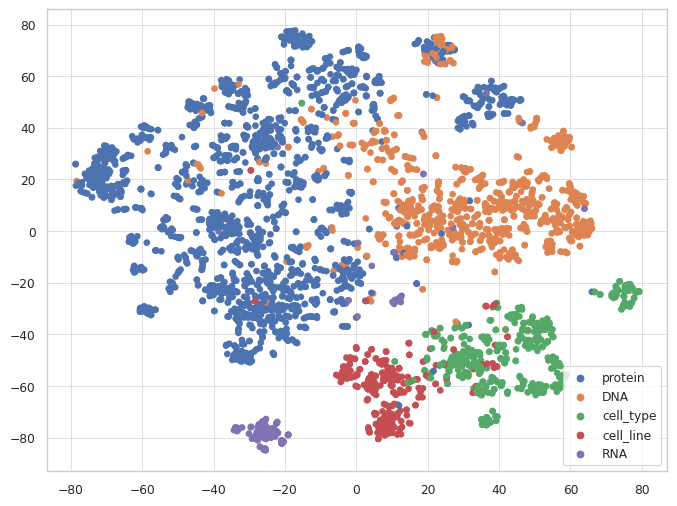}
        \caption{GENIA: Input to VIB}
        \label{fig:genia-beforeIB}
    \end{subfigure}
    \begin{subfigure}[b]{0.22\textwidth}
        \centering
        \includegraphics[width=\textwidth]{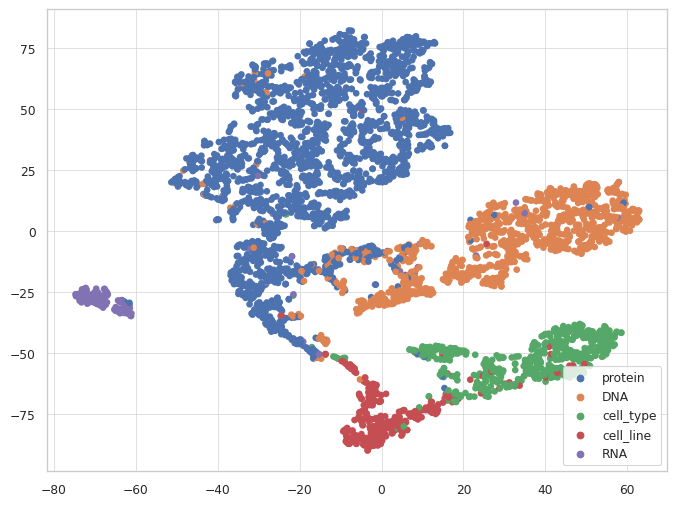}
        \caption{GENIA: Output by VIB}
        \label{fig:genia-afterIB}
    \end{subfigure}
    \begin{subfigure}[b]{0.22\textwidth}
         \centering
         \includegraphics[width=\textwidth]{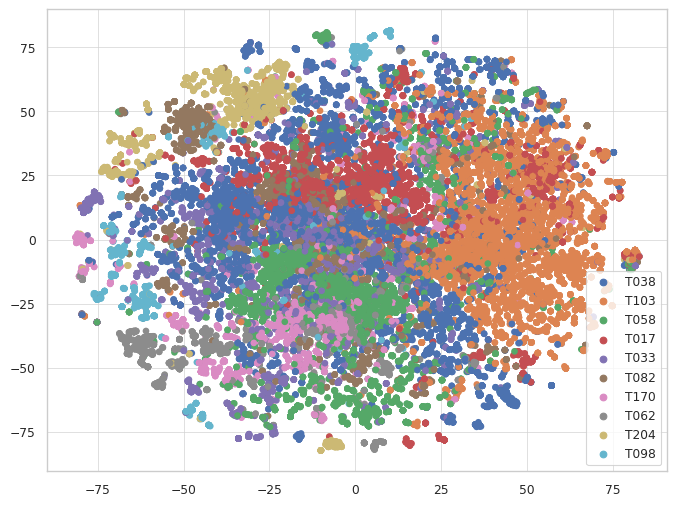}
         \caption{MM21: Input to VIB}
         \label{fig:mm21-beforeIB}
     \end{subfigure}
    \begin{subfigure}[b]{0.22\textwidth}
         \centering
         \includegraphics[width=\textwidth]{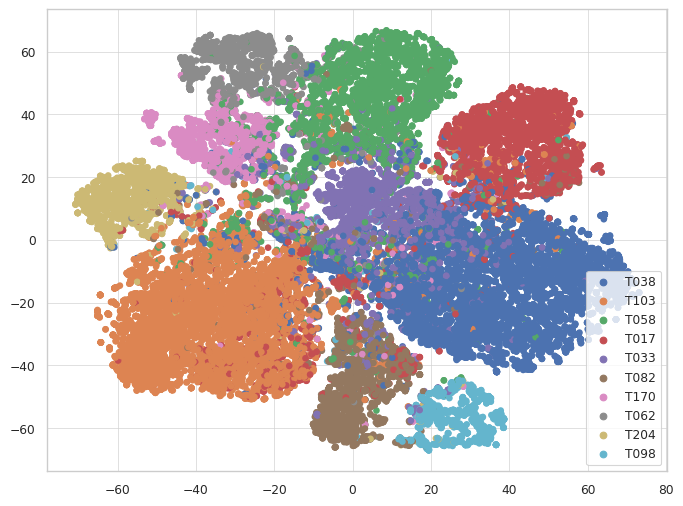}
         \caption{MM21-Output by VIB}
         \label{fig:mm21-afterIB}
     \end{subfigure}
    \caption{Visualisation of entity posteriors on the development set of GENIA and MedMention-21. Points in the same colour indicate entities in the same category.
    First row: posteriors from the generation models. Second row: the embeddings of entities input to VIB and the corresponding posteriors by VIB.
    }
    \label{fig:tsne}
\end{figure*}
\subsection{Entity Posteriors}
We further explore the effect of the proposed model by inspecting its intermediate values, which are posteriors $\bm{z}$ generated by the VAE and VIB components.

\subsubsection{Span Reconstruction and Synonym Generation}
We plot the resulting posteriors $\bm{z}_1$ of gold entities on the development set in a 2D space using tSNE~\cite{vanDerMaaten2008} in two cases: (i) Only VAEs: when we only train span reconstruction and synonym generation, and (ii) our joint model. 
We observed that across the settings, thanks to the target task of NER, $\bm{z}_1$ was satisfactorily restructured, i.e., posteriors of entities in the same category are clustered together. 
This phenomenon is more visible in the case of GENIA and MedMention-21\footnote{We only show embeddings of the top-10 named entity categories in MedMention-21.} than in the other corpora since the two corpora have more than one named entity categories, as illustrated in the first row of Figure~\ref{fig:tsne}.
In Figures~\ref{fig:genia-vae} and \ref{fig:mm21-vae}, entities are scattered over the space. Meanwhile, in Figures~\ref{fig:genia-joint} and \ref{fig:mm21-joint}, points in the same category are grouped. 
We can conclude that after simultaneously training VAEs and NER (the All model), we have better clusters than only training VAEs (Only VAEs).

\subsubsection{Supervised IB}
Similarly, we plot the gold entity embeddings and their posteriors $\bm{z}_3$ estimated by the supervised IB in the second row of Figure~\ref{fig:tsne}.
We can see in Figures~\ref{fig:genia-beforeIB} and \ref{fig:mm21-afterIB} that initially, the gold entity embeddings are not clustered very well.
In contrast, their posteriors are neatly grouped as illustrated in Figures~\ref{fig:genia-afterIB} and \ref{fig:mm21-afterIB}.
These clusters look even more distinguishable than those by $\bm{z}_1$, i.e., Figures~\ref{fig:genia-joint} and \ref{fig:mm21-joint}.
This phenomenon is understandable because $\bm{z}_3$ is approximated based on supervised learning while $\bm{z}_1$ is based on unsupervised one.

These visualisations also demonstrate the effectiveness of the supervised IB method.
While reducing the input representation size, the method can still reserve relevant features about the target classes.
As a result, we can have smaller but more meaningful representations than the original ones.
Such neat clusters are potentially helpful for entity linking~\cite{liu-etal-2021-self}.

\section{Conclusion}
We introduced a joint span-based NER model consisting of three components: VAE-based span reconstruction, VAE-based synonym generation, and VIB-based NER.
Each component plays a different role in learning span representation.
The VIB-based NER tries to preserve the information about the target NER category in the estimated latent while filtering out irrelevant information from the input.
The model is forced to keep contextualised span information when having the span reconstruction component.
Meanwhile, the synonym generation component indirectly injects semantic information about a span's synonyms.
When testing on five different corpora, we found that the joint model could perform better than the baseline.

The proposed model focuses on learning span representation, which is applicable not only to span-based NER but also to other span-based tasks such as event coreference resolution~\cite{Lu_Ng_2021} and question answering~\cite{li-choi-2020-transformers}. 
We plan to apply our model to such tasks in the future.

\section*{Limitations}
We find two main limitations of the proposed model.
Firstly, we need synonyms for each gold entity to train the synonym generation component. Unfortunately, this requirement is unsuitable for many NER corpora, such as CoNLL 2012 and ACE2005. 
One possible way to go around this is to treat the coreferences of an entity (if any) as synonyms. 
Secondly, similarly to some span-based NER models, our model suffers from a considerable number of all possible spans.
We can alleviate this limitation by detecting spans' boundary~\cite{Tan_Qiu_Chen_Wang_Huang_2020,Xu_Huang_Feng_Hu_2021} before classifying spans.
We, however, leave those as future work.


\section*{Acknowledgements}
We would like to thank Dr. Hai-Long Trieu for his help and fruitful comments on the initial implementation of the paper. This paper is based on results obtained from a project JPNP20006, commissioned by the New Energy and Industrial Technology Development Organization (NEDO).

\bibliography{anthology,custom}
\bibliographystyle{acl_natbib}

\appendix
\section{Hyper-parameter settings}
\label{apd:para} 
In Table~\ref{tab:param}, we show the hyper-parameter values used in our experiments.
In both SupVIB+SpanReco and All settings, we trained VAEs in 10 epochs before introducing NER into the models.
Our models were finetuned on NVIDIA Tesla V100 GPUs with 16GB of RAM using Optuna~\cite{Optuna}.
Depending on the size of the corpus, we ran different numbers of trials, which were in the range of [5,15].
\begin{table}[h]
\small
    \centering
    \begin{tabular}{l|c}
    \hline
    Hyper-parameter name & Value \\
    \hline
    $\beta$ & [1e-6, 1e-4]\\
    $\gamma$ & [1e-6, 1e-4]\\
    Latent size & [512, 768, 1024] \\
    Input size of the LSTM encoder & 200 \\
    Output size of the LSTM decoder & 256 \\
    NER learning rate & [1e-5, 3e-4] \\
    VAE learning rate & [1e-4, 1e-3] \\
    Batch size & 16\\
    Maximum length of a span & 14 \\
    Maximum sentence length & 512 \\
    Number of epochs & 20 \\
    Number of epochs to pretrain VAEs & 10 \\
    \hline
    \end{tabular}
    \caption{Hyper-parameter values}
    \label{tab:param}
\end{table}

\end{document}